\documentclass[letterpaper,twoside,10pt]{article}
\usepackage{amsmath,color,soulutf8,longtable,colortbl,setspace,xspace,url,pdflscape,amsfonts,amssymb}
\usepackage[square,numbers]{natbib}
\usepackage{latexsym}
\usepackage{graphicx}
\usepackage{mathrsfs}
\usepackage{geometry}
\usepackage{fancyhdr}
\usepackage{booktabs}
\usepackage{multirow}
\usepackage{array}
\usepackage[ruled]{algorithm}
\usepackage[font={footnotesize,bf,sf}]{caption}
\usepackage{xcolor}
\definecolor{gris}{rgb}{0.44,0.44,0.44}
\graphicspath{{Figures/}}
\PassOptionsToPackage{hyphens}{url}
\usepackage{hyperref} 
\hypersetup{colorlinks,%
	citecolor={black}, 
	urlcolor={blue},
	linkcolor={blue},
	breaklinks={true},
	pagebackref={true},
	hyperindex={true},
}

\makeatletter
\renewcommand{\section}{\@startsection {section}{1}{\z@}%
             {-2ex \@plus -1ex \@minus -.2ex}%
             {1ex \@plus.2ex}%
             {\normalfont\Large\sffamily\bfseries}}
\renewcommand{\subsection}{\@startsection{subsection}{2}{\z@}%
             {-1.25ex\@plus -1ex \@minus -.2ex}%
             {.75ex \@plus .2ex}%
             {\normalfont\large\sffamily\bfseries}}
\renewcommand{\subsubsection}{\@startsection{subsubsection}{3}%
             {\z@}%
             {-1.25ex\@plus -1ex \@minus -.2ex}%
             {.75ex \@plus .2ex}%
             {\normalfont\normalsize\sffamily\bfseries}}
\renewcommand\paragraph{\@startsection{paragraph}{4}{\z@}%
                                    {-1.25ex \@plus 1ex \@minus -.2ex}%
                                    {-.5em \@plus -.1em}%
                                    {\normalfont\normalsize\sffamily\bfseries}}
\def\@listI{\leftmargin\leftmargini    
            \parsep .25ex \@plus .1ex  
            \topsep .25ex \@plus .1ex  
            \itemsep \parsep}
\let\@listi\@listI
\@listi

\DeclareRobustCommand{\qed}{%
  \ifmmode \mathqed
  \else
    \leavevmode\unskip\penalty9999 \hbox{}\nobreak\hfill
    \quad\hbox{\qedsymbol}%
  \fi
}
\let\QED@stack\@empty
\let\qed@elt\relax
\newcommand{\pushQED}[1]{%
  \toks@{\qed@elt{#1}}\@temptokena\expandafter{\QED@stack}%
  \xdef\QED@stack{\the\toks@\the\@temptokena}%
}
\newcommand{\popQED}{%
  \begingroup\let\qed@elt\popQED@elt \QED@stack\relax\relax\endgroup
}
\def\popQED@elt#1#2\relax{#1\gdef\QED@stack{#2}}
\newcommand{\qedhere}{%
  \begingroup \let\mathqed\math@qedhere
    \let\qed@elt\setQED@elt \QED@stack\relax\relax \endgroup
}
\newcommand{\openbox}{\leavevmode
  \hbox to.77778em{%
  \hfil\vrule
  \vbox to.675em{\hrule width.6em\vfil\hrule}%
  \vrule\hfil}}
\DeclareRobustCommand{\textsquare}{%
  \begingroup \usefont{U}{msa}{m}{n}\thr@@\endgroup
}
\providecommand{\qedsymbol}{\openbox}

\providecommand{\proofname}{\sffamily Proof}

\makeatother

\allowdisplaybreaks

\def\numero{G--2019--26\ }
\def\mois{April 2019\ }
\def\titre{Flight-connection prediction for airline crew scheduling to construct initial clusters for OR optimizer}
\def\auteurs{Y.~Yaakoubi, F.~Soumis, S.~Lacoste-Julien}
\def\auteursC{Yaakoubi, Soumis, Lacoste-Julien}

\usepackage{breakcites}
\usepackage{tikz}
\usepackage{float}
\usetikzlibrary{calc, positioning, shapes}
\newcommand{\argmax}{\operatornamewithlimits{arg\,max}}
\DeclareRobustCommand{\1}[1]{\text{\usefont{U}{bbold}{m}{n}1}{\{#1\}}}
\newcolumntype{L}[1]{>{\raggedright\let\newline\\\arraybackslash\hspace{0pt}}m{#1}}
\newcolumntype{C}[1]{>{\centering\let\newline\\\arraybackslash\hspace{0pt}}m{#1}}
\newcolumntype{R}[1]{>{\raggedleft\let\newline\\\arraybackslash\hspace{0pt}}m{#1}}

\begin{document}
\def\figurename{{{\sffamily Figure}}} 
\def\tablename{{{\sffamily Table}}} 
\renewcommand{\thefigure}{{\sffamily \arabic{figure}}}
\renewcommand{\thetable}{{\sffamily \arabic{table}}}
\pagestyle{empty}

\begin{titlepage}
	\sffamily
	
{\noindent{\large\bfseries Les Cahiers du GERAD}\hfill ISSN:\quad
	0711--2440}
	
	\vspace*{7.5cm} \hspace*{240pt}
	\begin{minipage}[c][5.4cm][c]{7.9cm}
		{\begin{tabular}{p{.5cm}|p{6.5cm}}
				\multicolumn{2}{l}{\bfseries Flight-connection prediction for airline crew}\\
				\multicolumn{2}{l}{\bfseries scheduling to construct initial clusters}\\
				\multicolumn{2}{l}{\bfseries for OR optimizer}\\*[10pt]
				& Y.~Yaakoubi,  \\ 
				& F.~Soumis, S.~Lacoste-Julien\\*[12pt]
				&\numero\\*[8pt]
				&\mois
			\end{tabular}}
		\end{minipage}
		
		\vfill
		
\hrule
\smallskip

\noindent\begin{minipage}[t][][l]{7.5cm}
	\scriptsize
	La collection \textit{Les Cahiers du GERAD} est constitu\'{e}e des travaux de recherche men\'{e}s par nos membres. La plupart de ces documents de travail a \'{e}t\'{e} soumis \`{a} des revues avec comit\'{e} de r\'{e}vision. Lorsqu'un document est accept\'{e} et publi\'{e}, le pdf original est retir\'{e} si c'est n\'{e}cessaire et un lien vers l'article publi\'{e} est ajout\'{e}.\\ 
	
	\medskip
	\scriptsize
	\textbf{Citation sugg\'{e}r\'{e}e :} \auteurs~(Avril 2019). \titre, Rapport technique, Les Cahiers du GERAD G--2019--26, GERAD, HEC Montr\'{e}al, Canada.\\
	
	\textbf{Avant de citer ce rapport technique,} veuillez visiter notre site Web 
	(\url{https://www.gerad.ca/fr/papers/G-2019-26}) afin de mettre \`a jour vos donn\'ees de r\'ef\'erence, s'il a \'et\'e publi\'e dans une revue scientifique.\par
\end{minipage}
\hfill
\begin{minipage}[t][][l]{7.5cm}
	\scriptsize
	The series \textit{Les Cahiers du GERAD} consists of working papers carried out by our members. Most of these pre-prints have been submitted to peer-reviewed journals. When accepted and published, if necessary, the original pdf is removed and a link to the published article is added.\\ \\
	
	\scriptsize
	\textbf{Suggested citation:} \auteurs~(April 2019). \titre, Technical report, Les Cahiers du GERAD G--2019--26, GERAD, HEC Montr\'{e}al, Canada.\\
	
	\textbf{Before citing this technical report,} please visit our website (\url{https://www.gerad.ca/en/papers/G-2019-26}) to update your reference data, if it has been published in a scientific journal.\par
\end{minipage}

\bigskip


\hrule
\smallskip

\noindent\begin{minipage}[t][2.1cm][l]{7.5cm}
	\scriptsize
	La publication de ces rapports de recherche est rendue possible gr\^ace au soutien de HEC Montr\'eal, Polytechnique Montr\'eal, Universit\'e McGill, Universit\'e du Qu\'ebec \`a Montr\'eal, ainsi que du Fonds de recherche du Qu\'ebec -- Nature et technologies. 
	
	\medskip
	D\'ep\^ot l\'egal -- Biblioth\`eque et Archives nationales du Qu\'ebec, 2019\\
	\phantom{Depot legal} -- Biblioth\`eque et Archives Canada, 2019\par
\end{minipage}
\hfill
\begin{minipage}[t][2.1cm][l]{7.5cm}
	\scriptsize
	The publication of these research reports is made possible
	thanks to the support of HEC Montr\'eal, Polytechnique Montr\'eal, McGill University, Universit\'e du Qu\'ebec \`a Montr\'eal, as well as the Fonds de recherche du
	Qu\'ebec -- Nature et technologies. 
	
	\medskip
	Legal deposit -- Biblioth\`eque et Archives nationales du Qu\'ebec, 2019\\
	\phantom{Legal deposit} -- Library and Archives Canada, 2019\par
\end{minipage}

\hrule
\smallskip

\noindent
\begin{minipage}[t][1cm][l]{7.5cm}
	\begin{scriptsize}\raggedleft
		\textbf{GERAD} HEC Montr\'eal
		
		3000, chemin de la C\^ote-Sainte-Catherine
		
		Montr\'eal (Qu\'ebec) Canada H3T 2A7\par
	\end{scriptsize}
\end{minipage}
\hspace*{.35cm}\vrule\hfill
\begin{minipage}[t][1cm][l]{7.5cm}
	\begin{scriptsize}
		\textbf{T\'el.\,: 514 340-6053}
		
		T\'el\'ec.\,: 514 340-5665
		
		info@gerad.ca
		
		www.gerad.ca\par
	\end{scriptsize}
\end{minipage}

\bigskip
\hrule

\end{titlepage}

\newpage\cleardoublepage
\parindent=0pt
\sffamily
\vspace*{1cm}
	{\LARGE\bfseries \titre\par} 
\begin{minipage}[t][13cm][l]{7.5cm}	
	\vspace*{55pt}
	
	{\Large\bfseries Yassine Yaakoubi\,$^{a,c,d}$}\\*[8pt] 
	{\Large\bfseries Fran\c{c}ois Soumis\,$^{a,c}$}\\*[8pt] 
	{\Large\bfseries Simon Lacoste-Julien\,$^{b,d}$}\\*[24pt] 
	
	{\em $^a$~Department of Mathematics and Industrial Engineering, Polytechnique Montr\'eal (Qu\'ebec) Canada, H3C 3A7}\\[6pt] 
	{\em $^b$~Department of Computer Science and Operations Research, Universit\'e de Montr\'eal (Qu\'ebec) Canada, H3C 3J7}\\[6pt] 
	{\em $^c$~GERAD, Montr\'eal (Qu\'ebec), Canada, H3T 2A7}\\[6pt] 
	{\em $^d$~MILA, Montr\'eal (Qu\'ebec), Canada, H2S 3H1}\\[18pt]
	
	{\small\tt yassine.yaakoubi@polymtl.ca}\\
	{\small\tt francois.soumis@gerad.ca} \\
	{\small\tt slacoste@iro.umontreal.ca} \\ 

	\vfill
\end{minipage}
\vfill
{\bfseries \mois}\\*
{\bfseries Les Cahiers du GERAD}\\
{\bfseries \numero}\\
{\footnotesize Copyright \copyright\ 2019 GERAD, \auteursC}
\vspace*{0.05cm}

\hrule
\smallskip

\noindent
\begin{minipage}[t][][l]{7.5cm}
\begin{scriptsize}
	Les textes publi\'es dans la s\'erie des rapports de recherche \textit{Les Cahiers du GERAD} n'engagent que la responsabilit\'e de leurs auteurs. Les auteurs conservent leur droit d'auteur et leurs droits moraux sur leurs publications et les utilisateurs s'engagent \`{a} reconna\^{\i}tre et respecter les exigences l\'{e}gales associ\'{e}es \`{a} ces droits. Ainsi, les utilisateurs:
	\begin {itemize}
	\item Peuvent t\'{e}l\'{e}charger et imprimer une copie de toute publication du portail public aux fins d'\'{e}tude ou de recherche priv\'{e}e;
	\item Ne peuvent pas distribuer le mat\'{e}riel ou l'utiliser pour une activit\'{e} \`{a} but lucratif ou pour un gain commercial;
	\item Peuvent distribuer gratuitement l'URL identifiant la publication.
\end{itemize}
Si vous pensez que ce document enfreint le droit d'auteur, contactez-nous en fournissant des d\'{e}tails. Nous supprimerons imm\'{e}diatement l'acc\`{e}s au travail et enqu\^{e}terons sur votre demande.\par
\end{scriptsize}
\end{minipage}
\hfill
\begin{minipage}[t][][l]{7.5cm}
\begin{scriptsize}
The authors are exclusively responsible for the content of their research papers published in the series \textit{Les Cahiers du GERAD}. Copyright and moral rights for the publications are retained by the authors and the users must commit themselves to recognize and abide the legal requirements associated with these rights. Thus, users:
\begin{itemize}
	\item May download and print one copy of any publication from the public portal for the purpose of private study or research;
	\item May not further distribute the material or use it for any profit-making activity or commercial gain;
	\item May freely distribute the URL identifying the publication.
\end{itemize}
If you believe that this document breaches copyright please contact us providing details, and we will remove access to the work immediately and investigate your claim.\par
\end{scriptsize}
\end{minipage}

\thispagestyle{empty}
\parindent=15pt
\newpage

\setcounter{page}{2}
\renewcommand{\thepage}{\roman{page}}
\pagestyle{fancy}
\lhead[
\textcolor{gris}{\sffamily{\,}\thepage}\hfill
\textcolor{gris}{\sffamily \numero}\hfill 
\textcolor{gris}{\sffamily Les Cahiers du GERAD}
{\large\strut}\color{gris}{\hrule}
]
{
\textcolor{gris}{\sffamily{\,} Les Cahiers du GERAD}\hfill
\textcolor{gris}{\sffamily \numero}\hfill 
\textcolor{gris}{\sffamily\thepage}
{\large\strut}\color{gris}{\hrule}
}
\chead[]{}
\rhead[]{}
\lfoot[]{}
\cfoot[]{}
\rfoot[]{}

\renewcommand{\headrulewidth}{0pt}
\renewcommand{\footrulewidth}{0pt}
\rmfamily
\vspace*{5pt}

\paragraph{Abstract: }
We present a case study of using machine learning classification algorithms to initialize a large-scale commercial solver (GENCOL) based on column generation in the context of the airline crew pairing problem, where small savings of as little as 1\% translate to increasing annual revenue by dozens of millions of dollars in a large airline.
Under the imitation learning framework, we focus on the problem of predicting the next connecting flight of a crew, framed as a multiclass classification problem trained from historical data, and design an adapted neural network approach that achieves high accuracy (99.7\% overall or 82.5\% on harder instances). We demonstrate the usefulness of our approach by using simple heuristics to combine the flight-connection predictions to form initial crew-pairing clusters that can be fed in the GENCOL solver, yielding a 10x speed improvement and up to 0.2\% cost saving.


\vfill
\hrule
\smallskip
\paragraph{Acknowledgments: }
This work was supported by IVADO and a Collaborative Research and Development Grant from the Natural Sciences and Engineering Research Council of Canada (NSERC) and AD OPT, a division of Kronos Inc. The authors would like to thank these organizations for their support and confidence.

\newpage
\setcounter{page}{1}
\renewcommand{\thepage}{\arabic{page}}
\baselineskip=12.5pt 
\rmfamily

\section{Introduction}

As the global airline industry grows in size and volume, the complexity of airline scheduling problems increases significantly over time, despite the the advanced computational capacities available today.
%
In this work, we are interested in the crew pairing problem (CPP), one of the steps of crew scheduling problems. For each crew category and type of aircraft fleet, the CPP consists of finding a set of rotations (pairings) at minimum cost so that each active flight is carried out by a crew. When scheduling these pairings, additional constraints must also be met, which vary and are generally derived from work agreements of each airline.

\begin{figure}[H]
    \centering
    \resizebox{0.8\linewidth}{!}{
    \begin{tikzpicture}
\definecolor{lightblue}{HTML}{aed8e5}
\definecolor{dodgerblue}{HTML}{1b8ff9}
\definecolor{mypink}{HTML}{fcc2cd}
\tikzstyle{ellipsenode} = [ellipse, draw, minimum width=2cm, minimum height=1.2cm, align=center, very thick, fill=lightblue]
\tikzstyle{circlenode} = [circle, draw, minimum width=1cm, minimum height=1cm, align=center, very thick, fill=dodgerblue]
\tikzstyle{squarenode} = [rectangle, draw, minimum width=0.7cm, minimum height=0.7cm, align=center, very thick, fill=mypink]
\node[ellipsenode] (A) {Base};
\node[circlenode, right=0.65cm of A] (B1) {};
\node[circlenode, right=0.2cm of B1] (B2) {};
\node[circlenode, right=0.5cm of B2] (B3) {};
\node[squarenode, right=0.8cm of B3] (C) {};
\node[circlenode, right=0.8cm of C] (D1) {};
\node[circlenode, right=0.4cm of D1] (D2) {};
\node[ellipsenode, right=0.65cm of D2] (E) {Base};
\coordinate (L1) at ($(B1.north west)+(-0.25,0.25)$);
\coordinate (L2) at ($(B3.south east)+(0.25,-0.25)$);
\coordinate (L3) at ($(C.north west)+(-0.235,0.235)$);
\coordinate (L4) at ($(C.south east)+(0.235,-0.235)$);
\coordinate (L5) at ($(D1.north west)+(-0.25,0.25)$);
\coordinate (L6) at ($(D2.south east)+(0.25,-0.25)$);
\draw[densely dashed] (L1) rectangle (L2)node[midway,yshift=-0.9cm]{Duty period};
\draw[densely dashed] (L3) rectangle (L4)node[midway,yshift=-0.9cm]{Layover};
\draw[densely dashed] (L5) rectangle (L6)node[midway,yshift=-0.9cm]{Duty period};
\draw[very thick](A.east)--(B1.west);
\draw[very thick, red] (B1.east)--(B2.west);
\draw[very thick, red] (B2.east)--(B3.west);
\draw[very thick, red] (B3.east)--(D1.west);
\draw[very thick, red] (D1.east)--(D2.west);
\draw[very thick, -stealth, red] (D2.east)--(E.west);
\node[circlenode, below left=1cm and 2cm of E, minimum width=0.5cm, minimum height=0.5cm, text width=0cm] (F) {};
\node[squarenode, below=0.2cm of F,minimum width=0.5cm, minimum height=0.5cm, text width=0cm] (G) {};
\node[right, xshift=0.3cm] at (F) {Operated flight};
\node[right, xshift=0.3cm] at (G) {Layover};
\draw[red, ultra thick] ($(G.west)+(0,-0.8)$) -- ($(G.east)+(0,-0.8)$)node[right, xshift=0.02cm, black]{Flight-Connection};
\end{tikzpicture}}
    \caption[Illustration of a crew pairing.]{Illustration of a crew pairing. The flight-connection variable (in red) is defined to determine the next flight that a crew is going to perform, given an incoming flight.}
    \label{crew-pairing-fig}
\end{figure}
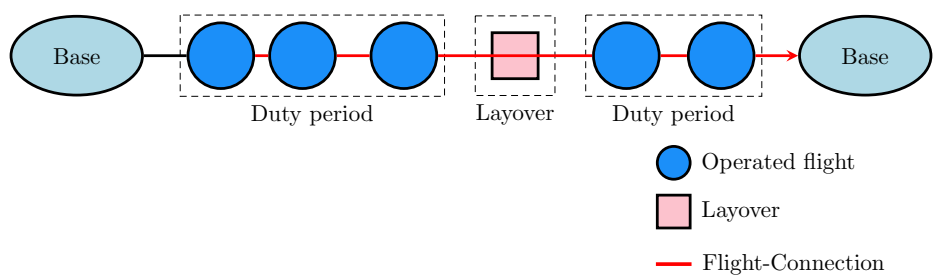

As in Figure \ref{crew-pairing-fig}, a pairing is a sequence of flights starting and finishing at a base, containing multiple duties (days of work) separated by a rest period.
For major airlines with more than 10,000 flights weekly, the CPP becomes an increasingly difficult problem to solve and efficient solutions are crucial since small savings of a mere 1\% translate into an increase of annual revenue by dozens of millions of dollars.
The complexity lies in the large number of possible pairings, and the selection of the set of pairings of minimal cost, which is a large integer programming problem impossible to solve with standard solvers~\citep{Elhallaoui2005}.

To reduce the number of variables and constraints to solve the CPP, many approaches have been developed and are used in current solvers to aggregate constraints to handle larger problems (see Section \ref{related}). However, these techniques require days to generate solutions, while airlines are often given all the scheduling data only a few days before having to build the schedule. Thus, the objective of this paper is to use machine learning (ML) techniques to improve algorithmic efficiency and solve the CPP in a more feasible time horizon.
While simple heuristics such as rule-based heuristics can be built to generate an initial solution, the cost of fine-tuning makes it obsolete in the long term, when new instances become available.
Our approach takes into consideration these instances and updates dynamically the weights of the neural network to propose an initial real-time solution without the need for manually fine-tuning the hyperparameters.
In addition, in contrast to problem-specific heuristics, our approach can be generalized to other problems as in airlines, bus, trucking, and rail industries.

Note that we are not aware of any prior work where ML is capable of solving a weekly CPP of this magnitude (10,000 flights).
Furthermore, learning from previous months' solutions, the structure of the 10,000 best \textbf{flight-connection variables}\footnote{The following flight that a crew is going to perform after each flight.} is a complex object to identify due to the large number of possible followers and the very complex costs and constraints on the sequence of flights in a pairing. Finding these variables with high accuracy for the solution would require more data than what is available.
Instead of using only ML (which requires more data than available) or the CPP solver (which takes tens of hours), we propose to use ML to obtain good initial information in a few seconds to start the optimizer, thus reducing the solution time by an order of magnitude. The information provided by ML does not need to be an entirely feasible solution~\citep{Elhallaoui2005}. Indeed, it can contain some weak points, which the CPP solver (GENCOL, referred to as OR optimizer) can fine-tune into a feasible solution. Naturally, the more the proposed initial solution presents weak points, the more the solver will take time to find a suitable solution. Thus, we are considering test accuracy as a primary metric to compare predictors in our paper.

\paragraph{Contributions.}
In order to obtain good initial information to speed up the CPP solver (GENCOL) that needs to be run on new schedule data, we exploit the large volume of flight data coming from the existing solution counting tens of thousands of flights over several months, and apply supervised learning algorithms to build a predictor for the \textbf{flight-connection} prediction problem.
We compare classical machine learning methods to neural networks and we adapt a neural network solution in several ways to improve the accuracy of the predictions: proposing improved feature encodings, reducing the number of classes to predict using domain-appropriate constraints, and testing solutions that can abstain to make a prediction (cf.\ Section~\ref{sec:algorithms}).

Using our algorithm, we can process tens of thousands of flights and propose good initial information in seconds instead of tens of hours, in contrast to current techniques reported in the literature.
As a proof of concept, we propose in Section~\ref{gencol-solver} simple heuristics to form initial crew pairing clusters for a weekly CPP that can be fed in the solver, yielding a 10x speed improvement and up to 0.2\% cost-saving, which would translate in millions of dollars saved for a large airline.

\section{Related Work}\label{related}

\subsection{Machine Learning for Scheduling }

Since many combinatorial optimization problems are NP-difficult, the use of ML techniques has been successfully proposed in several recent publications on various types of problems (e.g., in transportation, supply chain, energy, finance, and scheduling) \citep{Bengio2018}. The purpose of using ML is most often either to reduce the computational time of standard techniques to solve larger problems or to contribute to finding better solutions. For example, \citet{elmachtoub2017} propose the "Smart Predict-and-Optimize" where the authors present a framework to leverage the optimization problem structure and test on the shortest path problem (5$\times$5 grid) and portfolio selection problem (50 assets). \citet{mandi2020} use the same framework to solve more realistic discrete optimization problems: unweighted/weighted knapsack problem and energy-cost aware scheduling (up to 10 machines and 200 tasks). Note that this is different from the proposed approach in our paper, where the ML predictor does not have access to airline regulation, collective agreements, or cost function.

More generally, in cases where there is a structural understanding of the problem, ML can be used to provide quick approximations of the decisions to be made,  thus reducing the computation time required. This is known as the ``imitation learning'' \citep{Bengio2018}. Samples of the expected behavior are assumed to be available and are used in the ML model as demonstrations for learning. As such, the ML model has the task of minimizing a loss function by comparing the expert's decisions with his own. In other cases where one may not be able to understand the structure of the problem well enough, the ML model can be deployed to discover the structure itself. This is an application of reinforcement learning and we call this use case ``policy learning'' \citep{Bengio2018}. Since we have past CPP solutions, we use ML under the imitation learning framework.

This imitation learning framework has been used to solve multiple scheduling problems. For instance, \citet{Priore2014} provide an overview of the dynamic scheduling for manufacturing systems using ML techniques, using a set of previous system simulations (as training samples) in order to determine which rule is optimal for the current system state. Unlike the dynamic scheduling for manufacturing systems, we do not have access to these predefined constraints, as each airline company has its own collective agreement, and we would like to build a model that can be extended to \emph{other} airline companies. \citet{Wang2016} apply standard neural networks (NN) to predict the anticipated bus traffic, whereas \citet{Doherty2003} describes an application of NN to activity scheduling time horizon. Note that these approaches are not used in combination with an OR optimizer as in our case, and the flight-connection problem contains a larger number of classes as well as more complex constraints to satisfy.

Although many recent papers used ML to solve combinatorial problems similar to CPPs, to our knowledge, previous studies focused only on small-scale vehicle routing with up to 100 customers) \citep{nazari2018,kool2018} and airline crew scheduling problems with up to 714 flights \citep{deveci2018}. In our manuscript, we are interested in proposing an approach that can be scaled up to a large combinatorial problem with complex rules. While we test the approach on a weekly CPP with up to 10,000 flights, future research will focus on proposing a modified version of the solver in order to test the approach on a monthly CPP with up to 50,000 flights \citep{yaakoubi2020}. 

\subsection{Previous Work on Crew Pairing}\label{sec:prev_work_cp}

The CPP is known to be quite hard to solve \citep{Kasirzadeh2017}. Its complexity is due to the number of variables and constraints involved in the problem definition. In short, it is a large-scale integer programming problem. Although (meta)heuristics, such as simulated annealing \citep{Emden}, Tabu search \citep{Cavique}, genetic algorithms \citep{Levine,Ozdemir} and ant colony algorithms \citep{Deng} have been used in the literature to solve CPPs, they provide limited results. Instead, there are two common methods frequently employed: (i) branch-and-price (B\&P)~\citep{Freling2004,Gunluk2005,Brunner2013}; and (ii) Lagrangian relaxation \citep{Ceria1998,Caprara1999,Borenstein2009}.
The most prevalent method since the 1990s has been the set covering problem with column generation inserted in branch-\&-bound (see \citet{Desrochers1989}). This method, along with others, is described in a survey on CPPs by \citet{Cohn2003}; see also \citet{DEVECI201854} for a more recent survey. According to this latest survey, column generation was the most frequently used approach.

Column generation uses a master problem and a sub-problem. We refer to the master problem as the restricted master problem since we are using a subset of all possible pairings.
The sub-problem generates promising pairings that are more likely to improve the solution of the master problem. In the sub-problem, the constraints of flow continuity and local constraints ensure that flights in the pairing will be chained in time and space. Furthermore, rules and collective agreements have to be met; examples include minimum connection time between two flights, minimum rest time, and maximum number of duties in a pairing.

To reduce the number of constraints considered simultaneously, the work reported by \citet{Elhallaoui2005,Elhallaoui2008_multi} introduced a \emph{dynamic constraint aggregation} technique that decreases the amount of the set partitioning constraints in the restricted master problem and reduces the solution time.
Dynamic constraint aggregation (DCA) begins with an initial aggregation partition and aggregates in a single covering constraint each cluster of flights expected to be consecutive in an optimal solution. The algorithm modifies the clusters dynamically to reach an optimal solution if some expectations were wrong.
This modification strategy identifies the compatible and incompatible variables (pairings), where a variable (pairing) is said to be compatible with respect to the partition if the flights covered by the pairing correspond to a concatenation of some clusters. Otherwise, this variable is declared incompatible.
The partial pricing strategy used in this work uses the degree of incompatibility of a column, which is the number of times an incompatible column enters or exits in the middle of a cluster. This value can be computed in the sub-problem when a column is generated. The solver proceeds through a predetermined sequence of phases, typically, phases $k$ $=$ $0$, $1$, $2$, $\dots$ In phase $k$, only pairings with a degree of incompatibility not exceeding $k$ can be generated by the pricing problems.

Therefore, based on the algorithm's mechanism, two crucial difficulties could arise: (1) producing good initial clusters beyond using heuristic-rules based on the user's knowledge, and (2) preventing the algorithm from breaking connections in a cluster as much as possible. More specifically, we usually use a heuristic stopping criterion to control the solution time, the percentage of clusters to be broken during the resolution (e.g., 2\%), and the maximum phase to reach (e.g., phase 2).
As such, when using ML predictors to construct initial clusters of flights, we should determine which flight-connection variables are likely to be equal to one. Although these clusters may not necessarily represent a feasible solution, DCA will modify (repair) the solution with phase 1 and phase 2 of the simplex to reach a good feasible solution. Therefore, although there is no formal way to study the impact of test accuracy (of the ML predictor) on the objective function of the CPP, we do not break more than 1-2\% of the clusters. This justifies (1) using the test accuracy as a metric to compare ML predictors in Section~\ref{sec:experiments}, and (2) searching for the ``best'' predictor as opposed to sub-optimal ones for which the cost of the CPP solution deteriorates because the solver is not able to break all clusters containing mispredictions.

\section{Problem Setting} \label{Problem_setting}

\subsection{Crew Pairing Problem: Formulation and Motivation}
\label{crew-pairing motivation}

To solve the CPP, we provide the OR optimizer with an initialization which uses a partition of the flights into clusters. Each cluster represents a sequence of flights with a strong probability to be consecutive in the same pairing in the solution. 
To construct each cluster, we need to perform two seperate tasks. The first is to have information on where and when a crew begins a pairing, which makes it possible to identify whether a flight is the beginning of a pairing.
The second one is to predict, for each incoming flight to a connecting city, what the crew is going to do next: layover, flight, or end of a pairing. If it is the second case (flight), then we further predict which flight the crew will undertake.

Since the end of a pairing depends on the maximum number of days in a pairing permitted by the airline, we solely rely on this number as a hyperparameter
Therefore, to construct pairings, we propose to decompose predicting what the crew is going to do next into two sub-tasks. The first is predicting whether the crew makes a layover; the second is predicting the next flight, assuming the crew always takes another flight.
%
We refer to this problem as the \textbf{flight-connection problem}, which is the focus of our ML approach described in the next sections.

\subsection{Flight-connection Prediction Problem and Dataset} \label{flight-connection-dataset}

The flight-connection dataset contains several months of historical crew pairing data from an anonymous airline covering approximately 200 cities and tens of thousands of flights per month. We transform this data to build a flight-connection prediction problem (a multiclass classification problem) where the goal is to predict the next flight that an airline crew should follow in their schedule, given only partial information about the beginning of their schedule. To avoid error propagation through the whole sequence of flights (and because the OR solver is able to use partially correct plans), we only use information about the crew's \emph{incoming flight}
to predict the next flight.

The classification problem is thus the following: given the information about an incoming flight in a specific connecting city, choose among all the possible departing flights from this city (which can be restricted to the next 48 hours) the one that the crew should follow. These departing flights are identified by their flight code (about 2,000 possible flight codes). Different flights may share the same flight code in some airlines, as flights performed multiple times a week usually use the same flight code. Nevertheless, a flight is uniquely identified by its flight code, departing city and day; information which in practice can be deduced from the information about the incoming flight and a 48 hour-window.

Each flight is described using 5 features that are used in the classification algorithm: the origin and destination cities ($\sim$200 categories), the aircraft type (5 types), the flight duration (in minutes), and the time (with date) of arrival (for an incoming flight) or departure (for a departing flight).
Therefore, to predict the next flight that the crew takes, the input contains the information on the previous flight as well as the next 20 departing flights.

The dataset contains a train set and a test set.
The train set consists of 300,000 flight-connection variables, used for training and validation.
The test set is a benchmark used by airlines to decide which commercial solver to use and contains 10,000 flight-connection variables that are derived from a different weekly CPP~\footnote{The code and the dataset are available at: \url{ https://www.gerad.ca/fr/papers/G-2019-26 }}. Note that the importance of the methodology in this paper is illustrated in a follow-up paper of ours where we propose contributions to the OR solver, in order to use the proposed ML approach on a monthly CPP \citep{yaakoubi2020}.

\section{Algorithms} \label{sec:algorithms}


\subsection{Classical machine learning methods}\label{sec:classical-ML}

As neural networks have shown great potential to extract meaningful features from complex inputs and obtain good accuracy through different classification tasks, we propose to use them and compare results with classical ML methods: decision trees algorithms, Bayesian networks, logistic regression and ensemble methods.
Among the decision trees algorithms, we propose to use C4.5~\citep{Quinlan2014} and random forests~\citep{Breiman2001}.
As a probabilistic graphical model, Bayesian networks \citep{Pearl1982} provide a compact representation of a probability distribution that is too complex to be managed using traditional techniques.
Logistic regression uses multiple models, one model per class and use One-vs-rest strategy.
Ensemble methods (bagging, stacking and voting) generate and combine several classifiers to predict, thereby improving the generalization capability.

\subsection{Neural Network Model}

\begin{figure}
\centering
{
\resizebox{0.7\linewidth}{!}{
\resizebox{\textwidth}{!}{ 
\begin{tikzpicture}[font=\large]
\node[align=center,draw,anchor=west, rounded corners=2mm, minimum width=1cm, minimum height=1.2cm] (A) at (0,0) {Input \\ (2D)};
\node[align=center,draw,anchor=west, rounded corners=2mm, minimum width=1cm, minimum height=1.2cm] (B) at ([xshift=7mm]A.east) {Dense (120) \\ Relu activ. \\ \textbf{+} \\ Dropout};
\node[align=center,draw,anchor=west, rounded corners=2mm, minimum width=1cm, minimum height=1.2cm] (E) at ([xshift=7mm]B.east) {Dense (120) \\ Relu activ. \\ \textbf{+} \\ Dropout};
\node[align=center,draw,anchor=west, rounded corners=2mm, minimum width=1cm, minimum height=1.2cm] (F) at ([xshift=7mm]E.east) {Dense (120) \\ Relu activ. \\ \textbf{+} \\ Dropout};
\node[align=center,draw,anchor=west, rounded corners=2mm, minimum width=1cm, minimum height=1.2cm] (G) at ([xshift=7mm]F.east) {Dense (2,000) \\ Softmax activ.};

\draw[line width=1pt,-latex] (A.east)--(B.west);

\draw[line width=1pt,-latex] ([yshift=2.5mm]B.east)--([yshift=2.5mm]E.west);
\draw[line width=1pt,latex-, dotted] ([yshift=-2.5mm]B.east)--([yshift=-2.5mm]E.west);

\draw[line width=1pt,-latex] ([yshift=2.5mm]E.east)--([yshift=2.5mm]F.west);
\draw[line width=1pt,latex-, dotted] ([yshift=-2.5mm]E.east)--([yshift=-2.5mm]F.west);

\draw[line width=1pt,-latex] ([yshift=2.5mm]F.east)--([yshift=2.5mm]G.west);
\draw[line width=1pt,latex-, dotted] ([yshift=-2.5mm]F.east)--([yshift=-2.5mm]G.west);

\node[line width=0.5pt,draw=black!70,anchor=north west] at ([xshift=1cm,yshift=-8mm]A.south east) {\tikz[baseline=-1mm]\draw[line width=1pt,-latex]  (0pt,0pt) -- (10mm,0pt);  Forward flow \hspace{1cm} \tikz[baseline=-1mm]\draw[line width=1pt,-latex,dotted]  (10mm,0pt) -- (0pt,0pt);  Backward flow};
\end{tikzpicture}
}}}
\caption{The architecture diagram for the basic NN predictor. In this example, we use 3 hidden layers with 120 neurons in each. These values do not necessarily represent the best configuration of hyperparameters, which will be set using Bayesian Optimization.}
\label{fig:NN_arch}
\end{figure}
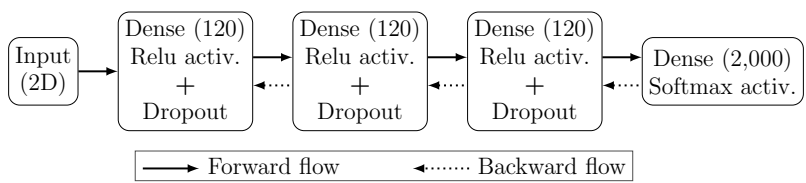

As in Figure \ref{fig:NN_arch}, we use a standard neural network with three dense hidden layers, each with 1,000 neurons and \texttt{relu} as the activation function. The output layer is a dense layer with 2,000 neurons since we predict the flight code and \texttt{softmax} as the activation function. We use dropout~\citep{srivastava14a} (with a probability of 0.2) to prevent over-fitting and batch normalization~\citep{batchnorm} between the activation function and the dropout. We refer to this predictor as the basic NN model.

Note that the city code and aircraft type features mentioned in Section \ref{flight-connection-dataset} are categorical features that are simply treated as numeric values in our basic NN model. This means that cities with nearby codes are treated similarly even though the codes are somewhat arbitrary.
To consider a more meaningful encoding for the different categorical features, we use one-hot encoding. We connect each one-hot encoding of a categorical feature to a separate hidden layer and we get an embedding layer of dimension $d$ for this feature ($d$ is a hyperparameter). This encoding (embedding layer) is learned during the NN training.
By concatenating the embedding layer's output for each categorical feature with the other numerical features (such as hours and minutes), we get an $n_d$-vector, where $n_{d}$ is the dimensionality of the representation of one flight that is fed into the NN. For the results given in Table~\ref{all-methods-results}, we feed the NN with the concatenation of the $n_d$ encodings of the incoming flight and the 20 next flights.

\subsection{Evaluation Metrics}\label{sec:evaluation_metrics}
We now present the evaluation metrics that we will consider.
Since the more weak points in the initial proposed solution the longer it will take the solver to find a suitable solution, we are particularly interested in the categorical accuracy and feasibility metric (motivating further refinements of the model described next).

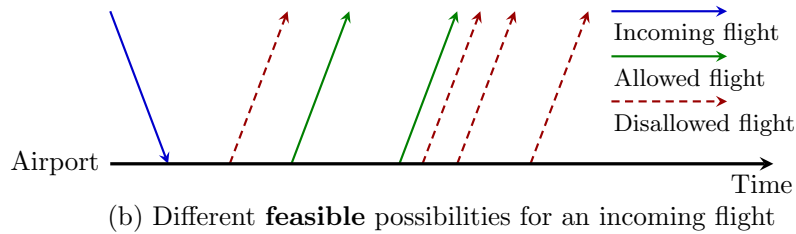
\begin{figure}[H]
    \centering
    \resizebox{0.7\linewidth}{!}{
    \begin{tikzpicture}[yscale=0.9]

\draw[thick, blue!80!black,-stealth] (0,2.2)--(0.75,0);
\draw[thick, red!60!black,-stealth, densely dashed] (1.55,0)--(2.3,2.2);
\draw[thick, green!50!black,-stealth, xshift=0.8cm] (1.55,0)--(2.3,2.2);
\draw[thick, green!50!black,-stealth, xshift=2.2cm] (1.55,0)--(2.3,2.2);
\draw[thick, red!60!black,-stealth, densely dashed, xshift=2.5cm] (1.55,0)--(2.3,2.2);
\draw[thick, red!60!black,-stealth, densely dashed, xshift=2.95cm] (1.55,0)--(2.3,2.2);
\draw[thick, red!60!black,-stealth, densely dashed, xshift=3.9cm] (1.55,0)--(2.3,2.2);
\draw[thick, blue!80!black,-stealth] (6.5,2.2)node[yshift=-0.3cm,xshift=-0.1cm, black,font=\small, anchor=west]{Incoming flight}--(8,2.2);
\draw[thick, green!50!black,-stealth, yshift=-0.65cm] (6.5,2.2)node[yshift=-0.3cm,xshift=-0.1cm, black,font=\small, anchor=west]{Allowed flight}--(8,2.2);
\draw[thick, red!60!black,-stealth, densely dashed, yshift=-1.3cm] (6.5,2.2)node[yshift=-0.3cm,xshift=-0.1cm, black,font=\small, anchor=west]{Disallowed flight}--(8,2.2);
\draw[very thick, -stealth] (0,0)node[left]{Airport}--(8.6,0)node[midway, yshift=-0.7cm]{(b) Different \textbf{feasible} possibilities for an incoming flight}node[below, xshift=-0.15cm]{Time};
\end{tikzpicture}}
    \caption{Illustration of an incoming flight scenario.}
    \label{incoming-flight-scenario}
\end{figure}

\begin{itemize}
\item Categorical accuracy:
$
\frac{1}{n} \sum\limits_{i=1}^n \1{ \argmax_{j}(y_{i,j}) ~ \in ~ \argmax_{j}(p_{i,j}) }
$,
where $i$ is the sample index, $j$ the class index, $y_{i,\cdot}$ the sample label encoded as a one-hot vector, and $p_{i,j}$ the classifier score for class $j$ for input $i$.

\item  top-$k$ categorical accuracy:
$
\frac{1}{n} \sum\limits_{i=1}^n \1{  \argmax_{j}(y_{i,j}) \in  \text{top-}k_{j}(p_{i,j}) } 
$.

\item Different aircraft accuracy:
Throughout the months, in 88\% to 95\% of the instances, the crew arrives at an airport and follows the aircraft, i.e.\ takes the next departing flight that the same aircraft makes (note that the classifier cannot know which next flight uses the same aircraft since we removed this information). To consider more difficult flight-connection instances, we report the accuracy only on the instances where the crew changes aircraft.

\item Feasibility: 
Given that the aircrew arrived at a specific airport at a given time, we can use \emph{a priori} knowledge to define which flights are possible. For example, as shown in Figure \ref{incoming-flight-scenario}, it is not possible to make a flight that starts ten minutes after arrival, nor is it possible five days later. Furthermore, it is rare that the type of aircraft changes between flights since each aircrew is formed to use one or two types of aircraft at most~\citep{Kasirzadeh2017}.
We use the following conditions that are always satisfied for the next flight taken by the crew:
\begin{itemize}
    \item The departure time of the next flight should follow the arrival time of the previous flight to the connecting city;
    \item The departure time of the next flight should not exceed 48 hours following the arrival time of the previous flight to the connecting city;
    \item The departure city of the next flight should be identical to the connecting city in the previous flight;
    \item The aircraft type should be the same throughout the same pairing since crew scheduling is separable by crew category and aircraft type~\citep{Kasirzadeh2017}.
\end{itemize}
We use these conditions to define a binary mask where $m_{i,j}=1$ only when flight $j$ satisfies these conditions for input $i$, and 0 otherwise. For the basic classifiers, we can evaluate the proportion of time they predict a feasible next flight with the ``feasibility'' metric:~$\frac{1}{n} \sum\limits_{i=1}^n  \1{  \argmax_{j}(p_{i,j}) \in  \argmax_{j}(m_{i,j}) } $\end{itemize}

\subsection{Masked Output Neural Network}\label{sec:masked}
To reduce the number of possible outputs, we can use the feasibility function $m_{i,j}$ to restrict our classifier to only predict a class among the feasible next flights for a specific incoming flight $i$.
We can use such a mask to define a new output layer $L_{m}$, such that in the output layer, for the softmax function, we only take into account the probabilities for the feasible next flights.

\subsection{Transformed Prediction Problem and Convolutional Neural Networks}

Using information on the feasibility of next flights, we can design a more useful encoding for output classes 
For each incoming flight, we consider all departing flights in the next 48 hours as a set. Then, we consider only those that are feasible according to our masking constraints (defined in Section \ref{sec:evaluation_metrics}) to filter this set. Then, we sort these flights based on the departure time, and we limit the maximum number of possible flights to 20, as it is sufficient in the airline industry. Thus, we predict the rank of the true label among that set.

\begin{figure}
{\centering
\resizebox{\textwidth}{!}{ 
\begin{tikzpicture}[font=\large]
\node[align=center,draw,anchor=west, rounded corners=2mm, minimum width=1cm, minimum height=1.2cm] (A) at (0,0) {Input \\ (2D)};
\node[align=center,draw,anchor=west, rounded corners=2mm, minimum width=1cm, minimum height=1.2cm] (B) at ([xshift=7mm]A.east) {Embedding \\ (3D)};
\node[align=center,draw,anchor=west, rounded corners=2mm, minimum width=1cm, minimum height=1.2cm] (C) at ([xshift=7mm]B.east) {3$\times$3 Conv (16) \\ \textbf{+} \\ 2$\times$2 Pooling};
\node[align=center,draw,anchor=west, rounded corners=2mm, minimum width=1cm, minimum height=1.2cm] (D) at ([xshift=7mm]C.east) {3$\times$3 Conv (16) \\ \textbf{+} \\ 2$\times$2 Pooling};
\node[align=center,draw,anchor=west, rounded corners=2mm, minimum width=1cm, minimum height=1.2cm] (E) at ([xshift=7mm]D.east) {Dense (120) \\ Relu activ. \\ \textbf{+} \\ Dropout};
\node[align=center,draw,anchor=west, rounded corners=2mm, minimum width=1cm, minimum height=1.2cm] (F) at ([xshift=7mm]E.east) {Dense (120) \\ Relu activ. \\ \textbf{+} \\ Dropout};
\node[align=center,draw,anchor=west, rounded corners=2mm, minimum width=1cm, minimum height=1.2cm] (G) at ([xshift=7mm]F.east) {Dense (20) \\ Softmax activ.};

\draw[line width=1pt,-latex] (A.east)--(B.west);

\draw[line width=1pt,-latex] ([yshift=2.5mm]B.east)--([yshift=2.5mm]C.west);
\draw[line width=1pt,latex-, dotted] ([yshift=-2.5mm]B.east)--([yshift=-2.5mm]C.west);

\draw[line width=1pt,-latex] ([yshift=2.5mm]C.east)--([yshift=2.5mm]D.west);
\draw[line width=1pt,latex-, dotted] ([yshift=-2.5mm]C.east)--([yshift=-2.5mm]D.west);

\draw[line width=1pt,-latex] ([yshift=2.5mm]D.east)--([yshift=2.5mm]E.west);
\draw[line width=1pt,latex-, dotted] ([yshift=-2.5mm]D.east)--([yshift=-2.5mm]E.west);

\draw[line width=1pt,-latex] ([yshift=2.5mm]E.east)--([yshift=2.5mm]F.west);
\draw[line width=1pt,latex-, dotted] ([yshift=-2.5mm]E.east)--([yshift=-2.5mm]F.west);

\draw[line width=1pt,-latex] ([yshift=2.5mm]F.east)--([yshift=2.5mm]G.west);
\draw[line width=1pt,latex-, dotted] ([yshift=-2.5mm]F.east)--([yshift=-2.5mm]G.west);

\node[line width=0.5pt,draw=black!70,anchor=north west] at ([xshift=3cm,yshift=-8mm]B.south east) {\tikz[baseline=-1mm]\draw[line width=1pt,-latex]  (0pt,0pt) -- (10mm,0pt);  Forward flow \hspace{1cm} \tikz[baseline=-1mm]\draw[line width=1pt,-latex,dotted]  (10mm,0pt) -- (0pt,0pt);  Backward flow};
\end{tikzpicture}
}}
\caption{The architecture diagram for the CNN predictor. In this example, we use 2 convolutional layers, 3 $\times$ 3 as filter size for the convolutional layers, 2 $\times$ 2 for pooling, 2 hidden layers with 120 neurons in each. These values do not necessarily represent the best configuration of hyperparameters, which will be set using Bayesian Optimization.}
\label{fig:CNN_arch}
\end{figure}
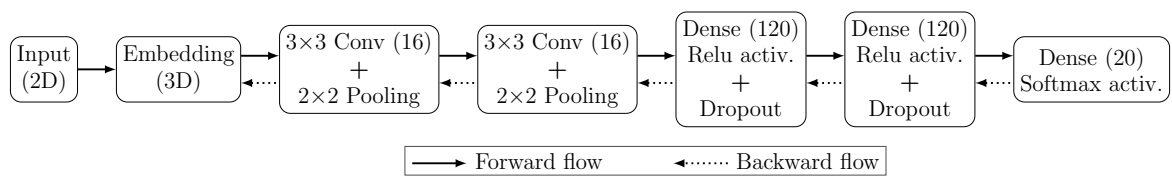

We use the embedding layer described earlier to construct a feature representation for each of these 20 flights. Then, we concatenate them to have a matrix $n_{d}\,\times\,$20 input for the next layers, where $n_{d}$ is the embedded representation of information on one flight. Consequently, we do not only reduce the number of possible next flights but also construct a similarity-based input, where the neighboring factors have similar features, which in turn allows the usage of convolutional neural networks (CNN), as in Figure \ref{fig:CNN_arch}. The intuition here is that we can consider each next flight as a different time step, enabling the use of convolutional architecture across time~\citep{Debayle2018,Borovykh2017}.

\subsection{Post-processing Step: Abstention Methods}
\label{abstention-section}

Since the OR solver uses phases starting with phase $0$, then $1$, then $2$, etc., it takes more time to break a connection (link) in a cluster than to build one.
In phase $k$, only pairings with a degree of incompatibility not exceeding $k$ can be generated by the pricing problems. Therefore, the solver first starts by combining the clusters given in the initial solution (phase $0$). Then, it starts to break these clusters at most once (phase $1$), then at most twice (phase $2$).
Thus, it becomes obvious to consider discarding a percentage of these predictions to enhance accuracy. Instead of taking all the predictions into account, removing low confidence links using this ``abstain'' option can be advantageous.
Therefore, we consider augmenting our classifier with an ``abstain'' option~\citep{nadeem10a}. Empirically, we observed that our solver is able to carry out the optimization problem much faster when taking into account $99\%$ of our predictions instead of $100\%$.

\subsubsection{Abstain when low confidence}\label{confidence-abstention}

We ignore the samples whose winning class confidence falls below a certain threshold. Indeed, \citet{Hendrycks} show that the lower the softmax probability of the most probable class, the greater the likelihood that the prediction is incorrect.
Therefore, we use a threshold for the probabilities given by the NN.
We also consider an RBF kernel SVM, to which we give as input the probabilities given by the NN, as well as the input of the NN. The binary label is $1$ when the NN predicted correctly, $0$ otherwise. Once trained, we put a threshold on the probabilities given by the SVM.

\subsubsection{Abstain when high entropy}

One issue with the approach of \citet{Hendrycks} is that modern neural networks are often notoriously miscalibrated~\citep{Guo17}. Methods such as temperature scaling~\citep{Guo17} can be applied to correct this miscalibration.
Unfortunately, even when we consider using calibration for selective classification, we still encounter problems in the presence of class imbalance (which is the case in the flight-connection dataset)~\citep{Fumera2000}.
The idea here, in short, is that we choose not to predict if the predictive entropy of the output probability vector $p$ (given by  $ H(p) = -\sum_i p_i \, \log p_i $) is too high~\citep{Hendrycks}.

\subsubsection{Estimate confidence with dropout}
The prediction is carried $N$ times while applying dropout in the entire layers of the NN~\citep{Gal2016}, yielding $N$ probability vectors for each test sample.
A rough estimate of certainty of prediction is obtained by computing the mean of these $N$ probability vectors and subtracting their component-wise estimated standard deviation (computed from the same $N$ vectors). This gives a rough lower bound on the certainty of our prediction. If the maximum value of these confidences is too low, we decide to abstain.

\section{Experiments} \label{sec:experiments}

\subsection{Hardware and Software}

The experiments were executed on a 40-core machine with 384\,GB of memory. Each method is executed in an asynchronously parallel set up of 2-4 GPUs.
All algorithms were implemented in Python using Keras~\citep{chollet2015keras} and Tensorflow~\citep{tensorflow2015} libraries.
For classical ML methods, we use Scikit-learn~\citep{scikit-learn}.

\subsection{Hyperparameter Selection}

\subsubsection{Random search}

As a first phase exploration of the hyperparameters space, we use random search~\citep{Bergstra2012} with k-fold cross-validation on the training set to choose the best hyperparameters for each method we consider. We use different months for different folds to simulate the more realistic scenario where we predict on a new time period. We keep the weekly problem of 10,000 flights for testing.

\subsubsection{Bayesian optimization}

After identifying the best predictor with random search, we refine further the choice of hyperparameters by using Bayesian optimization ~\citep{Hutter2011} with k-fold cross-validation to measure the configuration quality.
We optimize the hyperparameters listed in Table~\ref{hyper-parameters}~\citep{Dernoncourt2016} with an implementation of Gaussian process-based Bayesian optimization provided by the GPyOpt Python library~\citep{gpyopt2016}. The optimization is first initialized with 50 random search iterations, followed by 450 iterations of standard Gaussian process optimization.

\begin{table*}
\begin{center}
\caption{Hyperparameters used in optimization}
\resizebox{\linewidth}{!}{ 
\label{hyper-parameters}
\begin{tabular}{l|l|l}
\toprule
\textbf{Parameters} &
\textbf{Search space} &
\textbf{Type} \\
\midrule
Optimizer &
Adadelta; Adam; Adagrad; Rmsprop &
Categorical \\
Learning rate &
0.001, 0.002, \dots, 0.01 &
Float \\
Dimensions of the embeddings &
5, 10, 15, \dots, 50 &
Integer \\
Number of dense layers &
1, 2, 3, 4, 5 &
Integer \\
Neurons per layer &
100, 200, \dots, 1000 &
Integer \\
Dropout rate &
0.1, 0.2, \dots, 0.9 &
Float \\
Convolutional layers &
0, 1, 2, 3 &
Integer \\
Filters $n$ &
50, 100, 250, 500, 1000 &
Integer \\
Filter size h &
3,4,5 &
Integer \\
\bottomrule
\end{tabular}
}
\end{center}
\end{table*}

\subsection{Results}
\subsubsection{Classical machine learning methods}

Using the features described in Section \ref{flight-connection-dataset}, we consider several classical ML methods (see Section \ref{sec:classical-ML}).
As in Table \ref{table-classic}, since we find that C4.5 and random forests are highly more efficient than logistic regression and Bayesian networks, we have non-linear decision boundaries. In addition, since the train accuracy is greater than test accuracy, the predictors are over-fitting the data.
Ensemble methods yield better results than other predictors. Specifically, voting provides the best test accuracy (97.44\%).

Nonetheless, note that combining C4.5 and random forests presents several limitations. First, the hyperparameter configuration is less stable than it is for the NN, hence more fine-tuning is required. Second, these predictors have a very high memory usage. Although this can be limited by using aggressive pruning of the tree in C4.5 \citep{Quinlan2014}, this limits the performance of the predictor. Finally, they need to be retrained whenever there is a new instance (new unseen features: dates, cities, etc.). NN do not require such memory consumption, since we use batch-updates, and can be retrained with a lower learning rate on a new dataset.

\begin{table*}
\small
\begin{center}
\caption{Performance of classical machine learning algorithms}
\label{table-classic}
\resizebox{\linewidth}{!}{ 
\begin{tabular}{l|l|l|l|l}
\toprule
\textbf{Method} &
\textbf{Train accuracy (\%)} &
\textbf{Test accuracy (\%)} &
\textbf{Different aircraft accuracy (\%)} &
\textbf{Time (s)} \\
\midrule
Decision trees (C4.5) & 77.2 & 66.7 & 62.36 & 1000 \\
Logistic regression & 11.89 & 11.71 & 1.90 & 2000 \\
Bayesian networks & 6.61 & 6.32 & 0.62 & 3 \\
Random forests & 83.6 & 74.7 & 67.94 & 1000 \\
\midrule
Boosting (C4.5 and random forests) & 99.99 &  92.66 & 66.53 & 3000 \\
Stacking (C4.5 and random forests) & 97.4 & 87.8 & 69.26 & 3000 \\
Voting (C4.5 and random forests) & 99.99 &  97.44 & 61.84 & 2000 \\
\bottomrule
\end{tabular}
}
\end{center}
\end{table*}

\subsubsection{Neural networks}

After considering classical ML methods, we report results for NN with and without embedding as well as Masked output NN with embeddings (as described in section \ref{sec:masked}). We also report results for Transformed input with and without convolutional layers as well as Transformed input with convolutional layers after performing the Gaussian process (as described in section \ref{sec:gaussian-process}).
As in Table \ref{all-methods-results}, all NN predictors with embeddings outperform classical ML methods in term of test accuracy (99.35\%) and different aircraft accuracy (71.79\%). Furthermore, the computational effort (CPU training time) of executing the NN model without the embedding layer is much higher than other NN models, and provides a lower accuracy rate. When using the mask to guide the gradient descent during training jointly with the usage of the embedding layer, our model takes less than one-third of the training time and gives a better result put in comparison with the standalone embedding layer.

\begin{table*}
\small
\begin{center}
\caption{Comparison of results for different methods}
\label{all-methods-results}
\resizebox{\linewidth}{!}{ 
\begin{tabular}{l|l|l|l|l|l}
\toprule 
\textbf{Method} &
\textbf{Test} &
\textbf{Different aircraft} &
\textbf{Top-3} &
\textbf{Feasibility} &
\textbf{Time } \\
&
\textbf{accuracy (\%)}&
\textbf{accuracy (\%)}&
\textbf{accuracy (\%)}&
\textbf{(\%)}&
\textbf{(s)}\\
\midrule
NN without embeddings  &
88.03  &
60.52  &
98.95  &
92.25  &
4000 \\
NN with embeddings &
99.11 &
70.85 &
99.47 &
98.68 &
2000 \\
Masked output NN with embeddings  &
99.20  &
71.24  &
100  &
100  &
500 \\
Transf. input without convol. layers &
99.18 &
70.32  &
100  &
100 &
2500 \\
Transf. input with convol. layers &
99.35 &
71.79 &
100 &
100 &
700 \\
Transf. input with convol. layers (after GP) &
99.68 &
82.53 &
100 &
100 &
500 \\
\bottomrule
\end{tabular}
}
\end{center}
\end{table*}

\subsubsection{Gaussian process}
\label{sec:gaussian-process}

We perform the Gaussian process on the best predictor: ``Transformed input with convolutional layers'' to search for the best configuration of hyperparameters.
Using random search, after only a few iterations, we are able to get an accuracy of 99.35\%. Then, after 50 iterations, random search boosts the total return up to 99.62\% and we use Gaussian process to find the best architecture providing an accuracy of 99.68\%.
We can therefore conclude that the reported improvements (from classical ML methods to NN-based models) can be attributed to both the formulation of the classification problem as well as the NN architecture.
The final training was done with the best-identified architecture found by Bayesian optimization.

\subsubsection{Abstention}
\label{sec:abstention-results}

\begin{figure}
\centering
\includegraphics[width=0.55\linewidth]{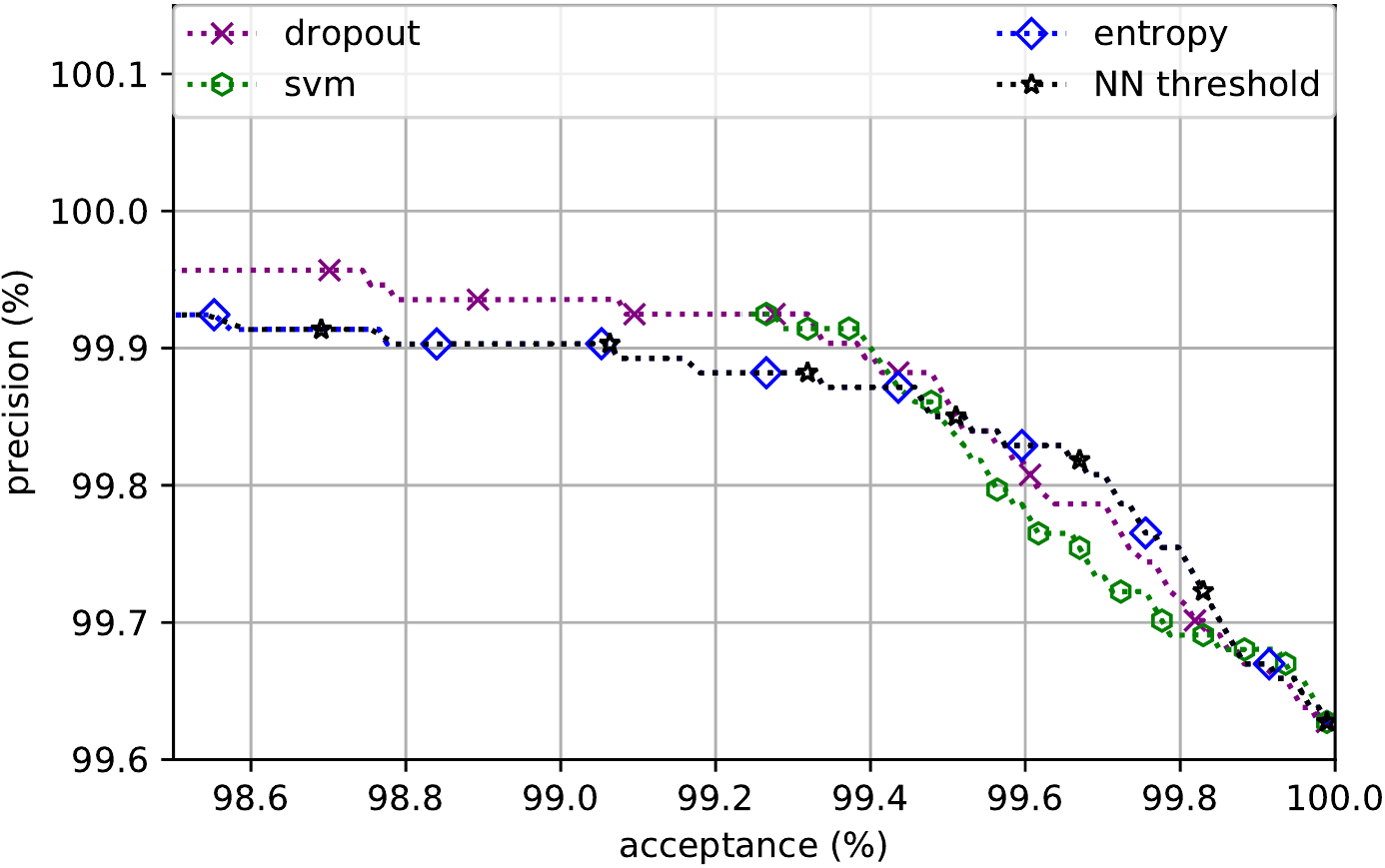}
\caption[Results of the different abstention methods]{Results of the different abstention methods.
NN threshold and Entropy use probabilities given by the NN.
SVM takes into consideration a confidence score given by a RBF-SVM.
Dropout uses the lower confidence bound of the prediction confidence.
For more information on these methods, see Section~\ref{abstention-section}}
\label{results-abstention}
\end{figure}

Figure \ref{results-abstention} shows the accuracy-abstention tradeoff curves for the abstention methods proposed in Section \ref{abstention-section} on the best architecture on the test set giving a 99.62\% accuracy initially; using these methods, we can discard some of our predictions and enhance the accuracy. For example, if we discard only 1\% of our predictions, the accuracy increases from 99.62\% to 99.90\% abstaining when low confidence and to 99.94\% estimating confidence with dropout.
For the next Section \ref{gencol-solver}, we will use a 1\% rejection rate.

\section{Integration of the Flight-Connection Predictor Into Crew Pairing} \label{gencol-solver}


GENCOL\footnote{\url{http://www.ad-opt.com/optimization/why-optimization/column-generation/}} (GENeration de COLonnes) is a commercial toolbox producing real-world airline crew scheduling solutions.
The toolbox uses column generation as an optimization strategy that allows a large number of variables to be considered in solving large-scale problems, with up to 3,000 flights.
We access this solver through private communication with the company.
As in section \ref{sec:prev_work_cp}, to solve larger problems, we use DCA (Dynamic Constraint Aggregation) to reduce the number of constraints by combining a few of them. This solver (GENCOL-DCA) starts with an initial aggregation partition, in clusters, of flights having a good probability of being done consecutively by the same crew, in an optimal solution.
As more than 20 airlines companies use GENCOL for their crew scheduling, we explore in this section whether our ML approach can yield gains on this state-of-the-art solver (through private communication with the company).

Upon the finalization of the flights-connection prediction model, we can use the same architecture to solve two other prediction problems on the test set (10,000 flights): (i) predict if each of the scheduled flights is the beginning of a pairing or not; and (ii) predict whether each flight is performed after a layover or not.
In reality, the three predictors share the same representation. To solve these independent classification problems, we sum the three prediction problems' cross-entropy losses when learning, therefore performing a multi-output classification.

We use a greedy heuristic to build the crew pairing. Specifically, we consider each flight that the model predicts at the beginning of a pairing as a first flight. Given this incoming flight, we predict whether the crew is making a layover or not. In both cases, we consider the incoming flight and predict the next one. The pairing ends when the crew has arrived at the base and when the maximal number of days permitted per pairing is approached.

We can use the above heuristic to construct a weekly solution for the testing data, obtaining a crew pairing that can be fed as an initial partition for the GENCOL-DCA solver.
Unfortunately, if one flight in the pairing is poorly predicted, as the flights are predefined, the crew can finish its pairing away from the base.
Therefore, we consider different heuristics to correct the pairings:
\begin{itemize}
\itemsep-0.25mm
    \item \textbf{Heuristic 1 (H1):} includes pairings where the crews finish away from the base, but it erases what occurs after the last time the crew arrives at the base, or deletes the whole pairing if it never passes to the base again.
    \item \textbf{Heuristic 2 (H2):} deletes pairings where crews finish away from the base;
    \item \textbf{Heuristic 3 (H3):} like heuristic 1, erases what occurs after the last time the crew arrives at the base but also cuts the pairings into smaller ones if we pass multiple times at the base.
\end{itemize}

As a baseline, we consider a benchmark called ``GENCOL init'', a standard initial solution approach obtained with the GENCOL solver.
The weekly problem is solved by a ``rolling time horizon'' approach with one-day windows and the week is divided into overlapping time slices of equal length (slightly more than one day).
Then a solution is constructed greedily in chronological order by solving the problem \emph{restricted} to each time slice sequentially, taking into account the solution of the previous slice through additional constraints.
We can feed the ``GENCOL init'' solution to the GENCOL-DCA solver as an initial partition for the constraint aggregation approach to further improve the solution globally, obtaining a solution that we call ``GENCOL-DCA from GENCOL init''. We can also use ML-based heuristics H1, H2, and H3 to construct clusters that we provide to the GENCOL-DCA solver as an initial partition, obtaining solutions that we respectively call ``GENCOL-DCA from H1'', ``GENCOL-DCA from H2'' and ``GENCOL-DCA from H3''. We report in Table \ref{table-solver} the savings w.r.t the cost of the ``GENCOL init'' solution and the total computational time.

\begin{table*}
\begin{center}
\caption[Final crew pairing costs]{Final crew pairing costs after running the GENCOL-DCA solver with different initialization clusters. We compare solution cost to cost of ``GENCOL init'' solution.}
\label{table-solver}
\resizebox{\columnwidth}{!}{ 
\begin{tabular}{l|l|l}
  \toprule
  \textbf{Resolution method} &
  \textbf{Cost diff. vs ``GENCOL init'' (\%)} &
  \textbf{Total time} \\
  \midrule
  GENCOL-DCA from GENCOL init & 1.21 & $\approx$ 45h \\
  & & \small{(40h + 5h)}\\
  \midrule
  GENCOL-DCA from H1 & \textbf{1,45} & 6h30 \\
  GENCOL-DCA from H2 & 1.38 & 5h \\
  GENCOL-DCA from H3 & 0.97 & 6h30\\
  \bottomrule
\end{tabular}
}
\end{center}
\end{table*}

As reported in Table \ref{table-solver}, two of the three heuristics that we propose outperform the benchmark solution ``GENCOL init'' as well as the ``GENCOL-DCA from GENCOL init'' solution, and we can conclude that for the pairings that finish away from the base, it is better to allow the solver to cover the flights than to propose smaller pairings.
Therefore, instead of performing the optimization process for 40 hours to obtain the ``GENCOL init'' solution and then for another 5 hours to obtain the ``GENCOL-DCA from GENCOL init'' solution, our proposed method gives better costs after a few seconds to predict the flight connections, and optimized results after five to six hours.

Note that, to our knowledge, there is no formal way to study the impact of test accuracy on the objective of the CPP. However, in these experiments, we observed that we did not break more than 2\% of the clusters (using a heuristic stopping criterion to control the resolution time). Therefore, although a different sub-optimal predictor could be used, if the test accuracy is below 98\% (as is the case for the classical ML methods and NN without embeddings), the GENCOL-DCA solver has to break more clusters to yield comparable results, which will increase resolution time. This is due to the fact that the GENCOL-DCA solver adds in the sub-problem a constraint forbidding in phase $k$ to generate a pairing breaking clusters more than $k$ times. During our experimentation, when using a sub-optimal predictor with the same stopping criterion, we observed that the cost of the solution is deteriorated because the solver is not able to break all clusters containing mispredictions.

Note that the heuristics that we propose to build crew pairings are still limited, even though they do give better costs in two of the three cases. Further study and experimentation  are required to explore other heuristics to construct the crew pairings.  Further study is also needed to solve \emph{monthly} problems by rolling horizon with one-week windows. The fast ML predictor will permit to construct in a few seconds clusters for each window of a week, customized according to the flight schedule of this week. We expect improvements in particular when the monthly schedule is irregular from week to week. It is the case nearly every month: Christmas, Easter, Thanksgiving, National Holiday, Mother and Father days, big sports events, etc. It was out of the question to use five times 40 hours to produce customized clusters for each window with the ``GENCOL init'' solution. Future studies will explore the generalization of the proposed approach to similar decision problems in different systems, such as bus, road, and rail traffic management.

\section{Conclusion}
It is rather difficult to assign the crew workers to a range of tasks while considering all the variables and constraints associated with the process. In this paper, we incorporate different algorithms and methods, which include neural networks in conjunction with abstention techniques, such as dropout, in order to prevent overfitting, or masks to ensure guidance of the gradient descent. These networks allow our solver to achieve high accuracy for the flight-connection problem.

We show that the usage of produced instances in weekly problems could lead to savings in cost. Consequently, the execution time may be enhanced drastically by varying the algorithm responsible for the pairings construction. This new algorithm will be used in conjunction with a modified OR solver in future work to permit an improved algorithmic efficiency and tests on monthly pairings.
Thus, our results are generalizable to larger problems (monthly CPPs with 50,000 flights) and to other domains (railway or bus-shift scheduling).

\bibliographystyle{ACM-Reference-Format}
\bibliography{G1926}

\end{document}